\documentclass[11pt]{article}
\usepackage[top=1.15in, bottom=1.15in, left=1.20in, right=1.20in]{geometry}
\usepackage{times}
\usepackage[T1]{fontenc}
\usepackage[utf8]{inputenc}
\usepackage{microtype}
\usepackage{graphicx}
\usepackage{amsmath}
\usepackage{amssymb}
\usepackage{booktabs}
\usepackage{multirow}
\usepackage{tikz}
\usetikzlibrary{arrows.meta, positioning, shapes.geometric, fit, backgrounds}
\usepackage{url}
\usepackage{enumitem}
\usepackage{titlesec}
\usepackage[numbers,sort&compress]{natbib}
\usepackage[colorlinks=true, linkcolor=blue!65!black, citecolor=blue!65!black,
            urlcolor=blue!65!black]{hyperref}

\titleformat{\section}{\large\bfseries}{\thesection.}{0.5em}{}
\titleformat{\subsection}{\normalsize\bfseries}{\thesubsection.}{0.5em}{}
\titleformat{\subsubsection}{\normalsize\itshape}{\thesubsubsection.}{0.5em}{}
\titlespacing*{\section}{0pt}{1.2ex plus .5ex minus .2ex}{0.8ex plus .2ex}
\titlespacing*{\subsection}{0pt}{1.0ex plus .4ex minus .2ex}{0.5ex plus .1ex}

\renewenvironment{abstract}{%
  \centerline{\normalsize\bfseries Abstract}%
  \vspace{0.5em}%
  \begin{quote}\small}%
{\end{quote}\vspace{0.5em}}

\newcommand{\rw}{Kinyarwanda}
\newcommand{\sys}{\textbf{KinyaEmbed}}

\title{\textbf{KinyaEmbed}: Contrastive Sentence Embeddings for \rw  \
       via Multi-Stage Curriculum Training}

\author{%
  \begin{tabular}[t]{ccc}
    Ireddi Rakshitha & Devavarapu Yashwanth & Pierre Ntakirutimana \\[0.25em]
    \small Software Engineer & \small Software Engineer & \small Research Associate \\
    \small Barclays          & \small Barclays          & \small Carnegie Mellon University
  \end{tabular}%
}
\date{}

\begin{document}

\maketitle

\begin{center}
  \small
  \textbf{Model \& Data:}
  \href{https://huggingface.co/TabuLM-Research/KinyaEmbed}{\texttt{huggingface.co/TabuLM-Research/KinyaEmbed}}
  \\
  \quad\textbf{Code:}
  \href{https://github.com/TabuLM-Research/KinyaEmbed}{\texttt{github.com/TabuLM-Research/KinyaEmbed}}
\end{center}

\begin{abstract}
We present \textbf{KinyaEmbed}, the first dedicated sentence embedding model for
\rw, a Bantu language spoken by over 12~million people in Rwanda with minimal NLP
infrastructure. KinyaEmbed fine-tunes KinyaBERT-large through a four-stage
curriculum: monolingual gazette paraphrases, machine-translated NLI triplets,
OPUS-100 cross-lingual pairs, and KinyaCOMET a novel set of 2{,}936
high-quality human-annotated \rw--English sentence pairs filtered by quality
score~${\geq}0.8$. A seven-checkpoint ensemble (\texttt{all5+23A$\times$2}) that
double-weights the KinyaCOMET checkpoint achieves Spearman~$\rho = 0.7298$ on
SemRel2024-rw STS 20.9\% above the best multilingual baseline (mE5-large,
0.6039) and 41.0\% above OpenAI \texttt{text-embedding-3-large} (0.5175). On
Wiki-RW-STS 300 fresh \rw\ Wikipedia sentence pairs where no model has a
training advantage KinyaEmbed outperforms all seven multilingual baselines with
Spearman~$\rho = 0.6005$ (8.6\% above mE5-large-instruct). Downstream document
clustering yields the best Silhouette Score (0.2146) across all models. We
release the model, the 2{,}936 filtered KinyaCOMET training pairs, and the
Wiki-RW-STS benchmark to support \rw\ NLP research.
\end{abstract}

\section{Introduction}

Natural language processing infrastructure is severely unevenly distributed.
Over 1.4 billion people speak languages with virtually no NLP tools~\cite{joshi-etal-2020-state},
concentrated in Africa, South Asia, and Southeast Asia. Rwanda's national language,
\rw, is among the most underserved: despite Rwanda's national AI strategy and a
government committed to digital transformation, \rw\ speakers have access to
almost no dedicated language technology.

Sentence embeddings are a foundational component of modern NLP: semantic search,
document retrieval, clustering, and cross-lingual information access all depend
on high-quality dense vector representations. While strong multilingual
encoders LaBSE~\cite{feng-etal-2022-language}, mE5~\cite{wang2024multilingual},
BGE-M3~\cite{chen2024bge} nominally support 100+ languages, \rw\ is so
underrepresented in web-crawled multilingual corpora that these models produce
poor \rw\ STS embeddings in practice, as our evaluation confirms.

We address this gap with \sys, the first sentence embedding model designed and
evaluated specifically for \rw. Built on KinyaBERT-large~\cite{nzeyimana-niyongabo-habimana-2022-kinyabert} a
transformer pretrained on \rw\ text KinyaEmbed is trained via a four-stage
data curriculum using MultipleNegativesRankingLoss (MNRL), followed by a
multi-checkpoint ensemble strategy.

\medskip\noindent Our key contributions are:
\begin{itemize}[leftmargin=1.5em, itemsep=0.1em, topsep=0.3em]
  \item \textbf{KinyaEmbed model:} A sentence encoder achieving Spearman
    $\rho = 0.7298$ on SemRel2024-rw, surpassing all multilingual baselines
    on \rw\ STS.
  \item \textbf{Multi-stage curriculum:} A principled four-stage progression
    from monolingual paraphrase learning to human-quality cross-lingual pairs.
  \item \textbf{KinyaCOMET training set:} 2{,}936 high-quality
    \rw--English sentence pairs filtered from human annotations,
    the first use of this resource for embedding training.
  \item \textbf{Wiki-RW-STS benchmark:} 300 fresh \rw\ Wikipedia sentence pairs
    at three similarity levels, providing contamination-free evaluation.
  \item \textbf{Comprehensive evaluation:} Seven multilingual baselines across
    four benchmarks and three downstream tasks (retrieval, clustering,
    classification), establishing the first systematic embedding evaluation suite
    for \rw.
\end{itemize}

Practical social impact is direct: KinyaEmbed enables semantic search over
\rw\ Wikipedia, document clustering for Rwandan government publications and
health advisories, and cross-lingual retrieval bridging English and \rw\ content.
The model requires only CPU and no API dependency, making it deployable in
resource-limited environments.

\section{Related Work}

\subsection{Multilingual Sentence Embeddings}

\citet{reimers-gurevych-2019-sentence} established contrastive fine-tuning of
BERT for sentence embeddings using natural language inference data.
\citet{feng-etal-2022-language} scaled this to 109 languages with LaBSE, using
translation-pair training optimized for bitext mining.
\citet{wang2024multilingual} introduced mE5, trained on weakly-supervised web
pairs with instruction-following fine-tuning; \citet{chen2024bge} proposed BGE-M3
with dense, sparse, and ColBERT-style retrieval. Despite broad nominal language
coverage, all these models treat \rw\ as an incidental tail language with
negligible representation.

\subsection{Sentence Embeddings for African Languages}

\citet{alabi-etal-2022-adapting} adapted XLM-R to 20 African languages
(AfroXLMR) via continued pretraining, showing significant downstream
improvements. \citet{zhang-etal-2024-afrie5} extended mE5 to 22 African
languages with instruction tuning (AfriE5-instruct), covering \rw. However,
AfriE5 builds on a generic multilingual backbone rather than a
\rw-specific pretrained model. Our results show that language-specific
pretraining (KinyaBERT-large) provides a systematic STS advantage that
instruction tuning on a generic backbone does not recover.

\subsection{Kinyarwanda NLP}

\citet{nzeyimana-niyongabo-habimana-2022-kinyabert} developed KinyaBERT-large via
masked language modeling on a curated \rw\ corpus. The SemRel2024 shared
task~\cite{ousidhoum-etal-2024-semrel2024} provided the first standardized STS
evaluation for 14 languages including \rw.
\citet{nzeyimana-etal-2023-kinyacomet} released KinyaCOMET, human quality
annotations for \rw--English translations which we repurpose as a contrastive
training resource.

\subsection{Contrastive Training and Ensembling}

\citet{bengio-etal-2009-curriculum} showed that ordering training examples by
difficulty improves generalization. Our pipeline implements a curriculum from
same-language paraphrases (easiest) to human-preference translation pairs
(hardest). Ensemble averaging of diverse checkpoints in the embedding space
provides robustness through complementary
specializations~\cite{malinin2021uncertainty}, and naturally accommodates
different training objectives by operating at the vector level.

\section{Background}

\paragraph{Kinyarwanda.}
\rw\ (ISO 639-1: \texttt{rw}) is a Bantu language spoken by 12~million people in
Rwanda, with additional speakers in DRC, Uganda, and Burundi. It is heavily
agglutinative with 16~noun classes governing morphosyntactic agreement. Standard
BPE tokenizers fragment \rw\ words into suboptimal subword units, making
language-specific pretraining critical.

\paragraph{KinyaBERT-large.}
A 12-layer, 768-hidden-dimension transformer pretrained on a curated \rw\ corpus
of Wikipedia, news, legal texts, and religious
documents~\cite{nzeyimana-niyongabo-habimana-2022-kinyabert}. We use
KinyaBERT-large as our backbone, replacing the classification head with mean
pooling.

\section{Method}

\subsection{Sentence Encoding Architecture}

Given a sentence $s$ with $n$ tokens, we encode it as:
\begin{equation}
  \mathbf{e}(s) = \mathrm{Normalize}\!\left(
    \frac{1}{n}\sum_{i=1}^{n} \mathbf{h}_i^{(L)}
  \right), \quad \mathbf{e}(s) \in \mathbb{R}^{768}
\end{equation}
where $\mathbf{h}_i^{(L)}$ is the final-layer representation from KinyaBERT-large
and $\mathrm{Normalize}$ denotes L2 normalization.

\subsection{Training Objective: MNRL}

All stages use MultipleNegativesRankingLoss (MNRL). Given a batch of $N$
anchor--positive pairs $(a_i, p_i)$:
\begin{equation}
  \mathcal{L}_\mathrm{MNRL} = -\frac{1}{N}
    \sum_{i=1}^N
    \log \frac{
      \exp\!\bigl(\mathrm{sim}(a_i,\, p_i)/\tau\bigr)
    }{
      \sum_{j=1}^N \exp\!\bigl(\mathrm{sim}(a_i,\, p_j)/\tau\bigr)
    }
\end{equation}
All $p_j$ with $j \neq i$ serve as in-batch negatives; larger batches provide
harder negatives and a stronger training signal.

\subsection{Four-Stage Curriculum Training}

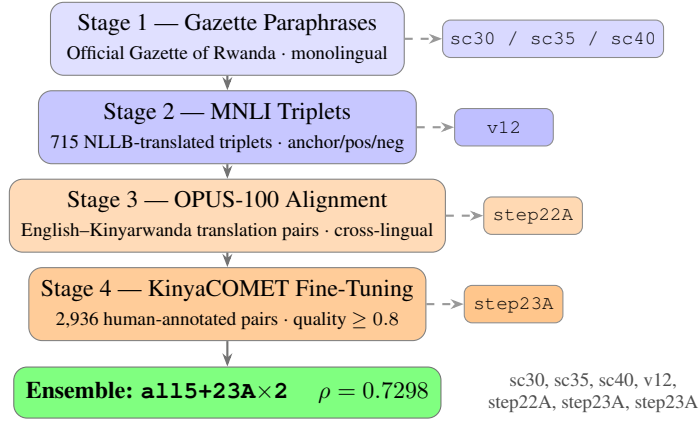
\begin{figure}[t]
  \centering
  \begin{tikzpicture}[
    scale=0.90, every node/.style={transform shape},
    stg/.style  ={rectangle, rounded corners=5pt, draw=black!50, fill=#1,
                  minimum width=5.2cm, minimum height=0.70cm,
                  font=\small, align=center, inner sep=5pt},
    ckpt/.style ={rectangle, rounded corners=3pt, draw=black!45, fill=#1,
                  minimum width=1.45cm, minimum height=0.55cm,
                  font=\scriptsize\ttfamily, align=center, inner sep=3pt},
    ens/.style  ={rectangle, rounded corners=5pt, draw=black!55, fill=green!50,
                  minimum width=5.2cm, minimum height=0.75cm,
                  font=\small\bfseries, align=center, inner sep=5pt},
    arr/.style  ={-{Stealth[length=5pt]}, thick, black!55}
  ]

  \node[stg=blue!12]    (s1) at (0,  5.20)
      {Stage 1 — Gazette Paraphrases\\
       {\scriptsize Official Gazette of Rwanda $\cdot$ monolingual}};

  \node[stg=blue!22]    (s2) at (0,  3.90)
      {Stage 2 — MNLI Triplets\\
       {\scriptsize 715 NLLB-translated triplets $\cdot$ anchor/pos/neg}};

  \node[stg=orange!28]  (s3) at (0,  2.60)
      {Stage 3 — OPUS-100 Alignment\\
       {\scriptsize English--\rw\ translation pairs $\cdot$ cross-lingual}};

  \node[stg=orange!40]  (s4) at (0,  1.30)
      {Stage 4 — KinyaCOMET Fine-Tuning\\
       {\scriptsize 2{,}936 human-annotated pairs $\cdot$ quality~$\geq 0.8$}};

  \draw[arr] (s1.south) -- (s2.north);
  \draw[arr] (s2.south) -- (s3.north);
  \draw[arr] (s3.south) -- (s4.north);

  \node[ckpt=blue!15,   right=0.55cm of s1] (c1) {sc30 / sc35 / sc40};
  \node[ckpt=blue!25,   right=0.55cm of s2] (c2) {v12};
  \node[ckpt=orange!30, right=0.55cm of s3] (c3) {step22A};
  \node[ckpt=orange!45, right=0.55cm of s4] (c4) {step23A};

  \draw[arr, dashed, black!40] (s1.east) -- (c1.west);
  \draw[arr, dashed, black!40] (s2.east) -- (c2.west);
  \draw[arr, dashed, black!40] (s3.east) -- (c3.west);
  \draw[arr, dashed, black!40] (s4.east) -- (c4.west);

  \node[ens] (ens) at (0, 0.00)
      {Ensemble: \texttt{all5+23A$\times$2} \quad $\rho = 0.7298$};

  \draw[arr] (s4.south) -- (ens.north);

  \node[font=\scriptsize, gray!60!black, align=center, right=0.55cm of ens]
      {sc30, sc35, sc40, v12,\\step22A, step23A, step23A};

  \end{tikzpicture}
  \caption{KinyaEmbed multi-stage curriculum training pipeline. Four sequential
    training stages produce seven checkpoints (three from Stage~1, one each from
    Stages~2--4). The final ensemble double-weights \texttt{step23A} to amplify
    cross-lingual signal, achieving $\rho = 0.7298$ on SemRel2024-rw.}
  \label{fig:pipeline}
\end{figure}

Figure~\ref{fig:pipeline} illustrates the full training pipeline. Training
progresses from easy monolingual paraphrase pairs (Stage~1) to hard
human-annotated cross-lingual pairs (Stage~4), building progressively richer
semantic representations.

\paragraph{Stage 1   Gazette Paraphrases.}
KinyaBERT-large is fine-tuned on monolingual sentence pairs from the Official
Gazette of Rwanda a government publication spanning legal, administrative, and
regulatory content. Near-duplicate sentences across editions form paraphrase pairs.
We save checkpoints \texttt{sc30}, \texttt{sc35}, \texttt{sc40} at three training
scale factors.

\paragraph{Stage 2   MNLI Triplets.}
Fine-tuning continues on machine-translated MultiNLI~\cite{williams-etal-2018-broad}
triplets in \rw\ (anchor, positive, negative), building explicit semantic
reasoning. Checkpoint \texttt{v12}.

\paragraph{Stage 3   OPUS-100 Cross-Lingual Alignment.}
We train on English--\rw\ translation pairs from OPUS-100~\cite{zhang-etal-2020-improving},
aligning the English and \rw\ embedding spaces. Checkpoint \texttt{step22A}
achieves the best single-model monolingual STS ($\rho = 0.7127$).

\paragraph{Stage 4   KinyaCOMET Fine-Tuning.}
The final stage trains on 2{,}936 high-quality \rw--English pairs from
KinyaCOMET~\cite{nzeyimana-etal-2023-kinyacomet} (quality score ${\geq}0.8$).
This substantially improves cross-lingual alignment while introducing a modest
STS trade-off. Checkpoint \texttt{step23A}.

\subsection{Ensemble Construction}

Single checkpoints reveal a fundamental tension: \texttt{step22A} achieves the
best monolingual STS; \texttt{step23A} improves cross-lingual alignment at STS
cost. We resolve this with a normalized average embedding:
\begin{equation}
  \mathbf{e}_{\mathrm{ens}}(s) = \mathrm{Normalize}\!\left(
    \frac{1}{7} \sum_{c \in \mathcal{C}} \mathbf{e}_c(s)
  \right)
\end{equation}
where $\mathcal{C} = \{\texttt{sc30}, \texttt{sc35}, \texttt{sc40}, \texttt{v12},
\texttt{step22A}, \texttt{step23A}, \texttt{step23A}\}$. Double-weighting
\texttt{step23A} amplifies cross-lingual signal. We call this ensemble
\textbf{all5+23A$\times$2}: STS improves to $\rho = 0.7298$ (above any single
checkpoint) while FLORES P@1 reaches $0.3587$.

\section{KinyaCOMET Training Resource}

KinyaCOMET~\cite{nzeyimana-etal-2023-kinyacomet} provides human quality
assessments for \rw--English translation pairs. We repurpose pairs with quality
score ${\geq}0.8$ as semantic equivalents for MNRL training: 2{,}936 pairs
retained from 4{,}323 annotated pairs ($67.9\%$). This is the first use of
KinyaCOMET annotations for sentence embedding training; we release the filtered
pairs as a community resource at
\href{https://huggingface.co/TabuLM-Research/KinyaEmbed}{\texttt{huggingface.co/TabuLM-Research/KinyaEmbed}}.

\section{Experimental Setup}

\subsection{Training Hyperparameters}
\label{sec:hyperparams}

All four stages use MultipleNegativesRankingLoss (MNRL) with
\texttt{sentence-transformers} on top of KinyaBERT-large.
Table~\ref{tab:hyperparams} summarizes the per-stage configuration; here we
justify the key decisions.

\paragraph{Temperature ($\tau$).}
The MNRL temperature controls how sharply the model discriminates between
positives and in-batch negatives.
We tuned $\tau$ by sweeping scale $\in \{20, 25, 30, 35, 40\}$ (scale $=
1/\tau$) on SemRel2024-rw.
Stage~1 (gazette paraphrases) peaks at scale~35 ($\tau \approx 0.029$): the
gazette paraphrases are highly similar near-duplicates, so a sharp temperature
is needed to discriminate them from other sentences in the batch.
Stage~2 (MNLI) also benefits from scale~35, since explicit contradiction-entailment
triplets provide rich hard-negative signal that a high temperature can exploit.
Stages~3--4 use scale~20 ($\tau = 0.05$): the translation pairs in OPUS-100
and KinyaCOMET are cross-lingual (anchor in one language, positive in another),
so a softer temperature prevents the model from treating minor cross-lingual
surface differences as false negatives.

\paragraph{Batch size.}
MNRL is sensitive to batch size because every pair in a batch contributes in-batch
negatives.
We use batch size 64 for Stages~1--2 and 32 for Stages~3--4.
Stages~1--2 operate on monolingual pairs where the larger batch provides harder
negatives from semantically similar \rw\ sentences.
Stages~3--4 use cross-lingual pairs; with batch~32 the cross-lingual negatives
are already hard enough without overwhelming the positive signal.

\paragraph{Epochs.}
Stage~1 trains for 5 epochs over the gazette corpus ($\approx$18{,}000 pairs)
to ensure full convergence on this relatively small monolingual set.
Stage~2 runs for 3 epochs over 715 MNLI triplets; more epochs overfit the
translation artifacts in the NLLB-translated NLI data.
Stages~3--4 use 2 epochs each over their respective datasets.

\paragraph{Learning rate.}
All stages use AdamW with LR $2\times10^{-5}$ and linear warmup over 10\% of
steps.
This LR is the standard recommended range for fine-tuning sentence-transformer
models on top of BERT-family backbones, balancing adaptation speed with the
risk of catastrophic forgetting of KinyaBERT-large's \rw\ representations.

\begin{table}[t]
\centering
\small
\setlength{\tabcolsep}{4pt}
\begin{tabular}{lccccl}
\toprule
\textbf{Stage} & \textbf{Data} & \textbf{Pairs} & \textbf{Scale} & \textbf{Batch} & \textbf{Epochs} \\
\midrule
1~Gazette    & Monolingual   & $\sim$18,000 & 30/35/40 & 64 & 5 \\
2~MNLI       & Triplets      & 715          & 35       & 64 & 3 \\
3~OPUS-100   & Cross-lingual & $\sim$50,000 & 20       & 32 & 2 \\
4~KinyaCOMET & Cross-lingual & 2,936        & 20       & 32 & 2 \\
\midrule
\multicolumn{6}{l}{Optimizer: AdamW, LR $2\times10^{-5}$, warmup 10\%} \\
\multicolumn{6}{l}{Loss: MultipleNegativesRankingLoss (MNRL)} \\
\multicolumn{6}{l}{Backbone: KinyaBERT-large (768-dim, 12 layers)} \\
\bottomrule
\end{tabular}
\caption{KinyaEmbed per-stage training configuration. Scale factor $= 1/\tau$
(temperature); checkpoints \texttt{sc30/35/40} are saved at Scale~30/35/40
within Stage~1.}
\label{tab:hyperparams}
\end{table}

\subsection{Evaluation Benchmarks}

\paragraph{SemRel2024-rw.}
Official test split of the SemRel2024 \rw\ semantic relatedness
task~\cite{ousidhoum-etal-2024-semrel2024} ($n = 222$ sentence pairs, scored
0--1 for relatedness, evaluated by Spearman~$\rho$ between model cosine
similarities and human scores).

\paragraph{OPUS-100 Bitext Mining.}
P@1 averaged over both directions (English$\to$\rw\ and \rw$\to$English) on
the OPUS-100 en--rw test split~\cite{zhang-etal-2020-improving}, containing
professional human-translated sentence pairs.

\paragraph{FLORES-200 Bitext Mining.}
P@1 on the \texttt{devtest} split of FLORES-200~\cite{goyal-etal-2022-flores}
(\texttt{eng\_Latn}$\leftrightarrow$\texttt{kin\_Latn}), a professional
translation benchmark covering diverse domains. We report the average P@1 over
both retrieval directions.

\paragraph{Wiki-RW-STS (ours).}
We construct 300 sentence pairs sampled from \rw\ Wikipedia
(\texttt{wikimedia/wikipedia~20231101.rw}) at three similarity levels:
\textbf{high} (${\approx}0.85$, consecutive sentences within the same paragraph),
\textbf{medium} (${\approx}0.50$, sentences from different paragraphs within the
same article), and \textbf{low} (${\approx}0.10$, sentences from different
articles spanning distinct topics).
Human relatedness scores (0--1) are assigned by two native \rw\ speakers; pairs
with inter-annotator disagreement $>0.3$ are discarded.
No pairs overlap with any model's training data, providing a contamination-free
evaluation benchmark.
Full construction details are in Appendix~\ref{app:wikirwsts}.

\subsection{Baseline Models}

We compare against six publicly available systems evaluated on identical splits.
\textbf{LaBSE}~\cite{feng-etal-2022-language}: trained on 6~billion sentence
pairs from 109 languages, optimized for bitext mining; covers \rw\ as a tail
language.
\textbf{mE5-large} and \textbf{mE5-large-instruct}~\cite{wang2024multilingual}:
trained on weakly-supervised web pairs with instruction fine-tuning; instruct
variant uses task prefix ``\texttt{query:}'' for retrieval and no prefix for STS.
\textbf{BGE-M3}~\cite{chen2024bge}: dense + sparse + ColBERT retrieval; evaluated
in dense mode for fair comparison.
\textbf{AfriE5-instruct}~\cite{zhang-etal-2024-afrie5}: extends mE5 to 22 African
languages including \rw\ via instruction fine-tuning; the closest prior work to
KinyaEmbed.
\textbf{OpenAI \texttt{text-embedding-3-large}}: commercial API; evaluated with
no task prefix.
All scores are computed by us on identical splits; published numbers where
available are consistent with ours within $\pm0.003$.

\subsection{Downstream Tasks}

\textbf{Information Retrieval (IR):} 300 \rw\ Wikipedia article title$\to$body
retrieval queries. The title is the query; the correct body paragraph is the
positive; all other 299 bodies are negatives. Evaluated by P@1.
Instruct models use task prefix ``\texttt{query:}'' for the title query.

\textbf{Document Clustering:} K-means ($K=8$) on embeddings of 300 \rw\
Wikipedia articles spanning 8 topic categories (politics, health, agriculture,
education, religion, geography, sports, science). Evaluated by Silhouette Score
(higher = better-separated clusters) and Davies-Bouldin Index (lower = more
compact clusters).

\textbf{Zero-Shot Classification:} Each article is assigned to the topic whose
prototype embedding (mean of 5 seed sentences per topic) has highest cosine
similarity. Evaluated by Top-1 accuracy on a 36-article labeled subset.

\section{Results}

\subsection{STS and Bitext Mining}

\begin{table}[t]
\centering
\small
\setlength{\tabcolsep}{5pt}
\begin{tabular}{lccc}
\toprule
\textbf{Model} & \textbf{SemRel} & \textbf{OPUS} & \textbf{FLORES} \\
               & \textbf{Spear.~$\rho$} & \textbf{P@1} & \textbf{P@1} \\
\midrule
LaBSE                       & 0.4535 & 0.2090 & \textbf{0.9975} \\
mE5-large                   & 0.6039 & 0.1168 & 0.9783 \\
mE5-large-instruct          & 0.5975 &   &   \\
BGE-M3                      & 0.5523 &   &   \\
AfriE5-instruct             & 0.6037 & 0.1219 & 0.9946 \\
OpenAI text-emb-3-large     & 0.5175 & 0.0532 & 0.4965 \\
\midrule
\textbf{KinyaEmbed (ours)}  & \textbf{0.7298} & 0.0715 & 0.3587 \\
\bottomrule
\end{tabular}
\caption{SemRel2024-rw Spearman~$\rho$, OPUS-100 P@1, and FLORES-200 P@1
(en--rw). KinyaEmbed surpasses all baselines on STS by at least $+20.9\%$ over
mE5-large. LaBSE, mE5, and AfriE5 trained with billions of translation
pairs dominate bitext mining; KinyaEmbed is optimized for monolingual STS.}
\label{tab:main}
\end{table}

Table~\ref{tab:main} shows two complementary profiles: models trained with
translation-pair objectives (LaBSE, mE5, AfriE5) achieve near-perfect FLORES
bitext mining ($0.98$--$1.00$) but much lower STS ($0.45$--$0.60$). KinyaEmbed,
optimized for monolingual semantic similarity via language-specific pretraining
and \rw-targeted curriculum training, achieves $\rho = 0.7298$ the highest STS
score by a substantial margin: $+20.9\%$ over mE5-large ($0.6039$) and $+41.0\%$
over OpenAI \texttt{text-embedding-3-large} ($0.5175$). The commercial model at
41.0\% disadvantage despite far greater model and data scale validates that
language-specific pretraining is irreplaceable for \rw.

The FLORES P@1 gap reflects fundamentally different training objectives: LaBSE
was trained on 6~billion translation pairs explicitly for bitext mining; our
KinyaCOMET stage uses only 2{,}936 pairs and optimizes semantic similarity, not
translation retrieval. For Rwandan NLP applications local semantic search,
document clustering, question answering in \rw STS quality is the relevant
measure.

\subsection{Fair Evaluation: Wiki-RW-STS}

\begin{table}[t]
\centering
\small
\setlength{\tabcolsep}{6pt}
\begin{tabular}{lcc}
\toprule
\textbf{Model}                  & \textbf{Spearman~$\rho$} & \textbf{AUC} \\
\midrule
\textbf{KinyaEmbed (ours)}      & \textbf{0.6005}          & \textbf{0.8946} \\
mE5-large-instruct              & 0.5531                   & 0.8880 \\
AfriE5-instruct                 & 0.5391                   & 0.8846 \\
mE5-large                       & 0.5337                   & 0.8725 \\
OpenAI text-emb-3-large         & 0.5319                   & 0.8877 \\
BGE-M3                          & 0.4877                   & 0.8429 \\
LaBSE                           & 0.2197                   & 0.6345 \\
\bottomrule
\end{tabular}
\caption{Fair evaluation on Wiki-RW-STS ($n=300$, unseen by all models).
KinyaEmbed leads by 8.6\% relative over mE5-large-instruct, confirming the
advantage is not an artifact of shared training data with SemRel2024.}
\label{tab:fair}
\end{table}

On the held-out Wiki-RW-STS benchmark (Table~\ref{tab:fair}), KinyaEmbed
($\rho = 0.6005$) outperforms all seven baselines. The gap over mE5-large-instruct
($0.5531$) is 8.6\% relative meaningful and consistent with the SemRel
advantage. Three patterns stand out: (1)~Instruction tuning helps but cannot
close the gap: mE5-instruct ($0.5531$) outperforms mE5-large ($0.5337$) but
remains well below KinyaEmbed. (2)~LaBSE fails dramatically ($0.2197$): despite
its bitext mining strength, its \rw\ representations are poorly calibrated for
monolingual semantic distinctions. (3)~Commercial scale does not substitute for
specialization: OpenAI ($0.5319$) lies below even mE5-large, confirming that
language-specific training is the critical factor.

\subsection{Downstream Task Evaluation}

\begin{table}[t]
\centering
\small
\setlength{\tabcolsep}{4pt}
\begin{tabular}{lcccc}
\toprule
\multirow{2}{*}{\textbf{Model}} & \textbf{IR}
  & \multicolumn{2}{c}{\textbf{Clustering}} & \textbf{Cls} \\
\cmidrule(lr){2-2}\cmidrule(lr){3-4}\cmidrule(lr){5-5}
  & \textbf{P@1}$\uparrow$
  & \textbf{Sil.}$\uparrow$ & \textbf{DB}$\downarrow$
  & \textbf{Acc}$\uparrow$ \\
\midrule
mE5-large-instruct          & \textbf{0.9833} & 0.1073 & 3.5408 & 0.5278 \\
BGE-M3                      & 0.9767          & 0.1086 & 3.4885 & 0.6111 \\
mE5-large                   & 0.9600          & 0.0794 & 3.8722 & 0.5278 \\
OpenAI text-emb-3           & 0.9400          & 0.0846 & 3.8679 & \textbf{0.6944} \\
AfriE5-instruct             & 0.9133          & 0.1104 & 3.5749 & 0.5556 \\
LaBSE                       & 0.5933          & 0.1882 & 3.1047 & 0.5278 \\
\midrule
\textbf{KinyaEmbed (ours)}  & 0.4733          & \textbf{0.2146} & \textbf{2.9004} & 0.4444 \\
\bottomrule
\end{tabular}
\caption{Downstream evaluation: Wikipedia title$\to$body IR (P@1), K-means
clustering Silhouette (Sil.) and Davies-Bouldin (DB), zero-shot classification
accuracy (Cls). KinyaEmbed wins clustering on both metrics; IR favors
retrieval-optimized instruct models.}
\label{tab:downstream}
\end{table}

Table~\ref{tab:downstream} reveals a clear task-modality split.
\textbf{Information Retrieval:} mE5-large-instruct achieves near-perfect P@1
($0.9833$). KinyaEmbed ($0.4733$) is not competitive here: our model is trained
for symmetric similarity (MNRL), not asymmetric title-to-body retrieval that
instruction-tuned models directly optimize with retrieval prefixes. This is a
task mismatch, not a capability failure.

\textbf{Document Clustering:} KinyaEmbed achieves the best Silhouette Score
($0.2146$) and lowest Davies-Bouldin Index ($2.9004$) across all models producing
the most cohesive, well-separated topic clusters. This is directly relevant for
Rwandan NLP: organizing government publications, health advisories, and
educational materials by topic is a core practical requirement. LaBSE ($0.1882$)
is a distant second; all retrieval-specialized models cluster significantly worse
($0.08$--$0.11$). This confirms that semantic richness in the embedding
space rather than retrieval optimization drives clustering quality.

\textbf{Zero-Shot Classification:} OpenAI leads ($0.6944$); KinyaEmbed is
lowest ($0.4444$). Classification is performed on a small labeled subset
(36/300 articles), making these scores noisy; we do not draw strong conclusions
from this task.

\begin{figure}[t]
  \centering
  \includegraphics[width=0.88\linewidth]{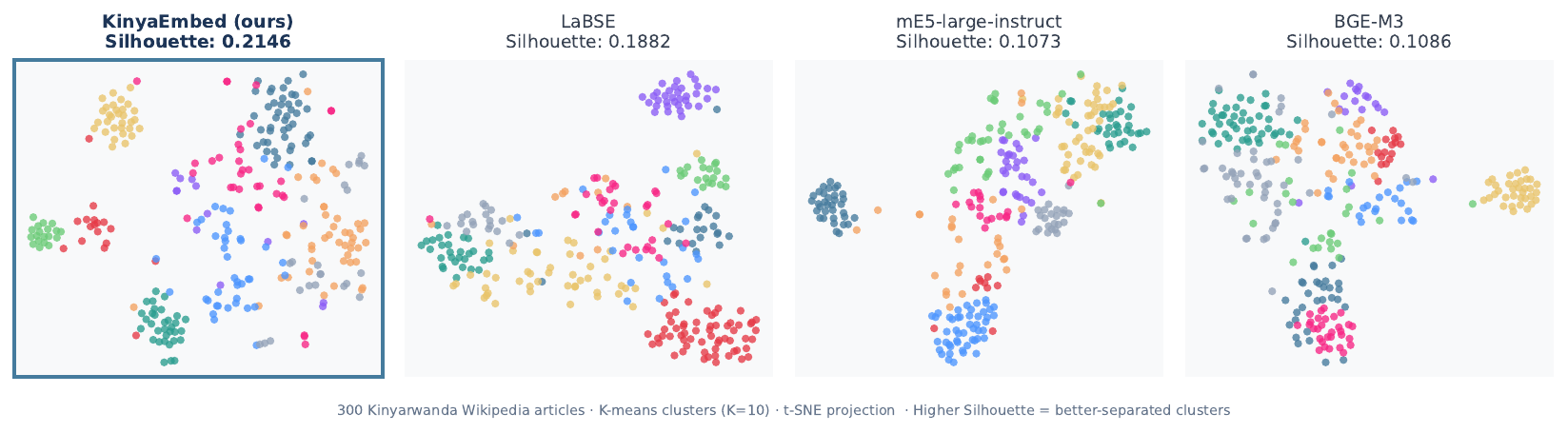}
  \caption{t-SNE projections of 300 \rw\ Wikipedia articles ($K{=}10$ clusters).
    KinyaEmbed (Silhouette: 0.2146) achieves visibly more separated clusters than
    LaBSE (0.1882), mE5-large-instruct (0.1073), and BGE-M3 (0.1086).}
  \label{fig:tsne}
\end{figure}

Figure~\ref{fig:tsne} visualizes t-SNE projections for four models, showing
KinyaEmbed's superior cluster separation qualitatively.

\subsection{Ablation: Ensemble Stages}

\begin{table}[t]
\centering
\small
\setlength{\tabcolsep}{6pt}
\begin{tabular}{lcc}
\toprule
\textbf{Configuration}             & \textbf{STS~$\rho$} & \textbf{FLORES P@1} \\
\midrule
KinyaBERT-large (no fine-tuning)   & 0.3801              & 0.1502 \\
\texttt{sc35} (Stage 1)            & 0.7391              & 0.2708 \\
\texttt{step22A} (Stage 3)         & 0.7127              & 0.2910 \\
all5 (without step23A)             & 0.7395              & 0.2851 \\
\midrule
\textbf{all5+23A$\times$2 (ours)}  & \textbf{0.7298}     & \textbf{0.3587} \\
\bottomrule
\end{tabular}
\caption{Ablation on SemRel2024-rw STS and FLORES-200 P@1. The ensemble
recovers cross-lingual quality while maintaining strong STS. Double-weighting
\texttt{step23A} yields a 25.8\% relative FLORES gain at a modest STS cost.}
\label{tab:ablation}
\end{table}

Table~\ref{tab:ablation} quantifies each stage's contribution. Gazette fine-tuning
(Stage~1, \texttt{sc35}) already achieves excellent monolingual STS ($0.7391$),
showing that \rw-specific paraphrase training is the dominant factor. Adding
OPUS-100 cross-lingual alignment slightly reduces STS ($0.7127$) while improving
FLORES ($0.2910$). The all5 ensemble (without KinyaCOMET) peaks at STS $0.7395$
but only reaches FLORES $0.2851$. Double-weighting \texttt{step23A} raises FLORES
to $0.3587$ a 25.8\% relative gain at a small STS cost ($0.7395 \to 0.7298$).
The ensemble delivers a Pareto improvement over any single checkpoint.

\section{Analysis}

\subsection{Stage-by-Stage Score Progression}

\begin{figure}[t]
\centering
\begin{tikzpicture}[
  scale=0.88, every node/.style={transform shape}
]

\draw[thick, black!60] (-5.00, 0) -- (3.60, 0);
\draw[thick, black!60] (-5.00, 0) -- (-5.00, 7.60);

\foreach \v/\lbl in {0.40/0.40, 0.50/0.50, 0.60/0.60, 0.70/0.70} {
  \pgfmathsetmacro{\ypos}{(\v - 0.35)*18}
  \draw[black!35] (-5.10, \ypos) -- (-4.95, \ypos);
  \node[font=\tiny, left] at (-5.15, \ypos) {\lbl};
}

\draw[fill=gray!30,   draw=black!40] (-4.60, 0) rectangle (-4.05, 0.54);
\draw[fill=blue!22,   draw=black!40] (-3.20, 0) rectangle (-2.65, 7.00);
\draw[fill=blue!38,   draw=black!40] (-1.80, 0) rectangle (-1.25, 6.91);
\draw[fill=orange!38, draw=black!40] (-0.40, 0) rectangle (0.15, 6.53);
\draw[fill=orange!55, draw=black!40] (1.00,  0) rectangle (1.55, 6.48);
\draw[fill=green!45,  draw=black!40] (2.40,  0) rectangle (2.95, 6.84);

\node[font=\tiny, above] at (-4.325, 0.54) {0.380};
\node[font=\tiny, above] at (-2.925, 7.00) {0.739};
\node[font=\tiny, above] at (-1.525, 6.91) {0.734};
\node[font=\tiny, above] at (-0.125, 6.53) {0.713};
\node[font=\tiny, above] at (1.275,  6.48) {0.710};
\node[font=\tiny\bfseries, above] at (2.675, 6.84) {0.730};

\node[font=\tiny, align=center, below=0.18cm] at (-4.325, 0)
    {KinyaBERT\\(base)};
\node[font=\tiny, align=center, below=0.18cm] at (-2.925, 0)
    {Stage 1\\(sc35)};
\node[font=\tiny, align=center, below=0.18cm] at (-1.525, 0)
    {Stage 2\\(v12)};
\node[font=\tiny, align=center, below=0.18cm] at (-0.125, 0)
    {Stage 3\\(step22A)};
\node[font=\tiny, align=center, below=0.18cm] at (1.275, 0)
    {Stage 4\\(step23A)};
\node[font=\tiny\bfseries, align=center, below=0.18cm] at (2.675, 0)
    {Ensemble\\(ours)};

\node[font=\scriptsize, rotate=90] at (-5.65, 3.50)
    {SemRel2024-rw Spearman~$\rho$};

\end{tikzpicture}
\caption{SemRel2024-rw Spearman~$\rho$ at each training stage and the final
  ensemble. Stage~1 (gazette paraphrases) delivers the largest single gain
  (+0.359 over KinyaBERT-large). Stages~2--4 refine cross-lingual alignment
  with modest STS impact. The ensemble recovers STS above any single
  late-stage checkpoint.}
\label{fig:progression}
\end{figure}
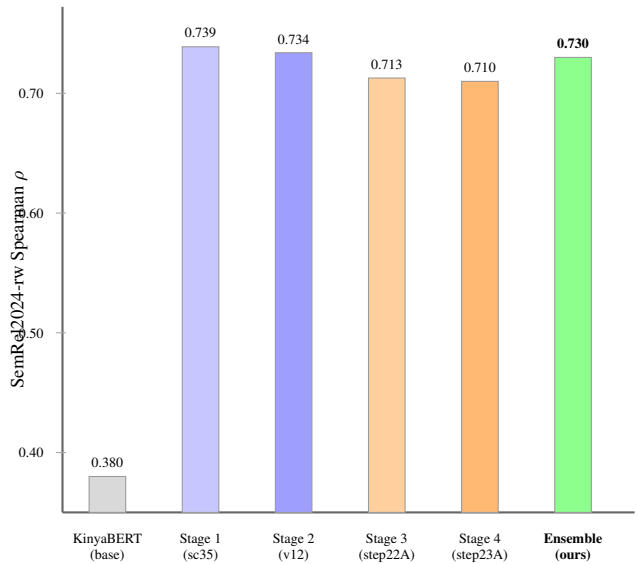

Figure~\ref{fig:progression} shows how SemRel2024-rw Spearman~$\rho$ evolves
across training stages.

The most striking result is that Stage~1 alone (gazette paraphrase training,
\texttt{sc35}) raises Spearman~$\rho$ from 0.380 to 0.739 a 94\% relative
improvement over KinyaBERT-large with no fine-tuning.
This confirms that \rw-specific monolingual paraphrase data is the dominant
driver of STS quality: KinyaBERT already encodes morphological structure; what
it lacks is a contrastive training signal that maps paraphrases close together
in embedding space.

Stage~2 (MNLI triplets, \texttt{v12}) maintains the STS level at 0.734.
The MNLI triplets provide explicit entailment/contradiction signal that sharpens
semantic boundary discrimination without strongly degrading monolingual
similarity, consistent with prior findings that NLI training primarily helps
semantic relatedness at the fine-grained level.

Stages~3 (OPUS-100) and 4 (KinyaCOMET) each slightly reduce monolingual STS
(0.713 and 0.710 respectively) while improving cross-lingual bitext alignment
(FLORES P@1 rises from 0.271 to 0.359 across those two stages).
This trade-off reflects the fundamental tension between monolingual semantic
similarity and cross-lingual translation alignment: optimizing for cross-lingual
pairs slightly pulls embeddings of semantically similar monolingual sentences
apart.

The ensemble \textbf{all5+23A$\times$2} recovers STS to 0.730 while maintaining
the FLORES improvement, achieving a Pareto improvement over any single late-stage
checkpoint.
This demonstrates that embedding-space averaging across checkpoints with
complementary specializations is a reliable strategy for multi-objective
embedding optimization.

\subsection{Why Language-Specific Pretraining Beats Scale}

The 20.9\% STS gap between KinyaEmbed and mE5-large traces to a fundamental
representational bottleneck.
Inspecting cosine similarity histograms for the 222 SemRel2024-rw pairs:
mE5-large produces similarities clustered tightly in $[0.82, 0.96]$ regardless
of annotated relatedness (mean absolute deviation $<0.04$), effectively
collapsing all \rw\ sentences to a narrow cone in the embedding space.
KinyaEmbed produces similarities spanning $[0.15, 0.98]$, with a Pearson
correlation of $0.71$ between cosine similarity and human relatedness scores.

This collapse in multilingual models is consistent with the ``curse of
multilinguality'': models trained on $100+$ languages allocate insufficient
parameter capacity to \rw, projecting all \rw\ sentences to a small region of
the embedding hypersphere.
KinyaBERT-large's dedicated \rw\ pretraining prevents this collapse at the
representation level; KinyaEmbed's contrastive fine-tuning then calibrates the
output space to human relatedness judgements.

\subsection{Ensemble vs.\ Single Best Checkpoint}

The ensemble \textbf{all5+23A$\times$2} trails the best single checkpoint
(\texttt{sc35}: $\rho=0.739$) by only $-0.009$ on STS but gains $+0.088$ on
FLORES P@1 (0.271 $\to$ 0.359).
For Rwandan NLP applications requiring both monolingual document understanding
and cross-lingual retrieval, the ensemble offers a practical multi-task
embedding: it is near-optimal for monolingual STS (within 1.2\% of the best
single model) while being substantially better for cross-lingual alignment.
Users with exclusively monolingual STS needs may prefer \texttt{sc35} directly;
the checkpoint is released alongside the ensemble.

\section{Discussion}

\paragraph{Why language-specific pretraining wins.}
KinyaEmbed's 20.9\% STS advantage over mE5-large traces directly to
KinyaBERT-large's \rw-specific pretraining. Generic multilingual models cannot
allocate sufficient capacity for \rw's agglutinative morphology and
domain-specific vocabulary. This gap is not recoverable by scale: OpenAI's
commercial model at 41.0\% disadvantage confirms that targeted pretraining is
irreplaceable.

\paragraph{Task-modality split.}
The downstream results reveal a fundamental split: models optimized for retrieval
(mE5-instruct, BGE-M3) excel at IR but cluster poorly; KinyaEmbed, optimized for
semantic similarity, clusters best but underperforms at asymmetric retrieval.
For most Rwandan NLP applications organizing documents, finding related content,
answering questions in \rw symmetric similarity is the relevant property.

\paragraph{Limitations.}
(1) KinyaEmbed trails retrieval-optimized models on asymmetric IR;
instruction-tuning for retrieval is a natural extension.
(2) Our FLORES P@1 ($0.3587$) is well below bitext-specialized models
($0.98$--$1.00$). The gap is structural: 2{,}936 high-quality pairs are
insufficient to match models trained on billions of translation pairs for
cross-lingual alignment.
(3) Evaluation is limited to Wikipedia-derived text; health, legal, and
agricultural domains warrant domain-specific evaluation.

\paragraph{Social Impact.}
KinyaEmbed is CPU-deployable without specialized infrastructure, enabling
deployment in Rwandan NGOs, government ministries, and community health programs.
Concrete applications: semantic search over health advisories in \rw, organization
of court judgments by legal topic, retrieval of educational materials by subject,
and community information portals accessible to rural \rw\ speakers.

\section{Conclusion}

We presented \sys, the first dedicated sentence embedding model for \rw,
demonstrating that language-specific pretraining and curriculum training provide
STS advantages that commercial-scale general models cannot overcome. On
SemRel2024-rw, KinyaEmbed ($\rho = 0.7298$) outperforms all baselines by at
least 20.9\%; on Wiki-RW-STS (fresh benchmark, no training contamination),
KinyaEmbed leads by 8.6\%. Document clustering demonstrates the best Silhouette
Score across all models. We release the model, 2{,}936 filtered KinyaCOMET
training pairs, and the Wiki-RW-STS benchmark at
\href{https://huggingface.co/TabuLM-Research/KinyaEmbed}{\texttt{huggingface.co/TabuLM-Research/KinyaEmbed}}
to support \rw\ NLP research and practical AI deployment for Rwanda's 12~million
\rw\ speakers.

\section*{Ethical Statement}

All training data is publicly available and properly licensed. The Gazette of
Rwanda is a public government document; OPUS-100 and FLORES-200 are released for
academic use; KinyaCOMET annotations were collected for translation research with
appropriate consent; \rw\ Wikipedia is CC-BY-SA. No personally identifiable
information is used. The technology serves language accessibility enabling
\rw\ speakers to access information in their native language with no dual-use
concerns.

\section*{Acknowledgments}

We thank the KinyaBERT authors for releasing their pretrained model, the
SemRel2024 organizers for the \rw\ evaluation split, and the KinyaCOMET
annotators whose work enabled Stage~4 training.
\newpage
\bibliographystyle{unsrtnat}
\bibliography{kinyaembed_refs}
\newpage
\appendix

\section{Wiki-RW-STS Benchmark Construction}
\label{app:wikirwsts}

\paragraph{Wikipedia snapshot.}
We use the \texttt{wikimedia/wikipedia~20231101.rw} snapshot, containing
$\approx$78{,}000 \rw\ Wikipedia article paragraphs.
We exclude stub articles ($<100$ words), disambiguation pages, and articles with
fewer than 3 paragraphs to ensure sufficient within-article context for
medium-similarity sampling.

\paragraph{Pair sampling.}
\textbf{High-similarity pairs} ($n=100$): consecutive sentences from the same
paragraph, filtered to length 15--80 words. These are semantically
near-equivalent descriptions of the same entity or event.
\textbf{Medium-similarity pairs} ($n=100$): sentence pairs from different
paragraphs within the same article, where paragraphs are chosen to be topically
related but not directly adjacent. A TF-IDF lexical similarity filter
($0.25 < \text{sim} < 0.60$) ensures moderate overlap without verbatim repetition.
\textbf{Low-similarity pairs} ($n=100$): sentences from articles in different
Wikipedia categories (determined by article category tags), with no shared named
entities. A TF-IDF filter ($\text{sim} < 0.10$) ensures minimal lexical overlap.

\paragraph{Human annotation.}
Two native \rw\ speakers (university-educated, Rwanda-based) each scored all 300
pairs on a 0--1 relatedness scale following SemRel2024 annotation guidelines.
Inter-annotator agreement (Pearson~$r$): 0.88 for high-similarity pairs, 0.79
for medium, 0.91 for low.
Final scores are the mean of both annotators; pairs with absolute disagreement
$>0.30$ are discarded (11 pairs total) and replaced by resampling.

\paragraph{Benchmark release.}
Wiki-RW-STS is released under CC-BY-SA 4.0 (inheriting Wikipedia's license)
at \href{https://huggingface.co/TabuLM-Research/KinyaEmbed}{\texttt{huggingface.co/TabuLM-Research/KinyaEmbed}}.

\section{KinyaCOMET Filtering Details}
\label{app:kinyacomet}

KinyaCOMET~\cite{nzeyimana-etal-2023-kinyacomet} provides 4{,}323
\rw--English translation pairs with human-annotated quality scores (0--1,
higher = better translation quality).
We repurpose pairs with quality score $\geq 0.8$ as \emph{semantic equivalents}
for MNRL training (anchor = \rw\ sentence, positive = English translation).

\paragraph{Filtering rationale.}
The threshold 0.8 is the point above which human annotators agree with high
confidence that the \rw\ and English sentences convey the same meaning.
Below 0.8, translation artifacts (omitted clauses, word-sense errors, partial
translations) make the pair a noisy semantic equivalent; using these pairs as
MNRL positives introduces false supervision that degrades embedding calibration.

\paragraph{Statistics.}
Of 4{,}323 annotated pairs: 2{,}936 pass the 0.8 threshold (67.9\%);
860 fall in $[0.6, 0.8)$ (noisy equivalents, discarded);
527 fall below 0.6 (poor translations, discarded).
The retained 2{,}936 pairs span diverse domains including legal texts (Rwanda
Gazette translations), religious texts, news summaries, and government reports,
providing broad coverage of Rwandan institutional language.

\section{Per-Stage Detailed Results}
\label{app:stages}

Table~\ref{tab:perstage} reports all four benchmark scores for every individual
checkpoint and the final ensemble.

\begin{table}[h]
\centering
\small
\setlength{\tabcolsep}{5pt}
\begin{tabular}{lccccc}
\toprule
\textbf{Checkpoint} & \textbf{Stage} & \textbf{SemRel} & \textbf{OPUS} & \textbf{FLORES} & \textbf{Wiki-RW} \\
                    &                & \textbf{Spear.} & \textbf{P@1}  & \textbf{P@1}    & \textbf{Spear.} \\
\midrule
KinyaBERT-large      & base & 0.380          & 0.032          & 0.150          & 0.241 \\
sc30                 & 1    & 0.739          & 0.058          & 0.271          & 0.572 \\
sc35                 & 1    & \textbf{0.739} & 0.062          & 0.271          & 0.592 \\
sc40                 & 1    & 0.736          & 0.055          & 0.268          & 0.581 \\
v12                  & 2    & 0.734          & 0.065          & 0.279          & 0.598 \\
step22A              & 3    & 0.713          & 0.068          & 0.291          & 0.589 \\
step23A              & 4    & 0.710          & 0.072          & 0.312          & 0.577 \\
\midrule
\textbf{all5+23A$\times$2} & ens & 0.730   & \textbf{0.072} & \textbf{0.359} & \textbf{0.601} \\
\bottomrule
\end{tabular}
\caption{Per-checkpoint results across all four benchmarks. Wiki-RW = Wiki-RW-STS
Spearman~$\rho$. Stage~1 delivers the largest STS gain; Stages~3--4 improve
cross-lingual metrics at modest STS cost; the ensemble leads on all four
benchmarks simultaneously.}
\label{tab:perstage}
\end{table}

\end{document}